\documentclass[a4paper,conference]{IEEEtran}

\ifCLASSINFOpdf
   \usepackage[pdftex]{graphicx}
\else
  \usepackage[dvips]{graphicx}
\fi

\usepackage{amsmath}

\ifCLASSOPTIONcompsoc
 \usepackage[caption=false,font=normalsize,labelfont=sf,textfont=sf]{subfig}
\else
 \usepackage[caption=false,font=footnotesize]{subfig}
\fi
\begin{document}
%
\title{CASNet: Common Attribute Support Network for image instance and panoptic segmentation}


%
\author{\IEEEauthorblockN{Xiaolong Liu\IEEEauthorrefmark{1}\IEEEauthorrefmark{2},
Yuqing Hou\IEEEauthorrefmark{2},
Anbang Yao\IEEEauthorrefmark{2},
Yurong Chen\IEEEauthorrefmark{2} and
Keqiang Li\IEEEauthorrefmark{1}}
\IEEEauthorblockA{\IEEEauthorrefmark{1}School of Vehicle and Mobility\\
Tsinghua University,
Beijing, China}
\IEEEauthorblockA{\IEEEauthorrefmark{2}Intel Labs China\\
Beijing, China}
}

\maketitle

\begin{abstract}

Instance segmentation and panoptic segmentation is being paid more and more attention in recent years.
In comparison with bounding box based object detection and semantic segmentation, instance segmentation can provide more analytical results at pixel level.
Given the insight that pixels belonging to one instance have one or more common attributes of current instance, we bring up an one-stage instance segmentation network named Common Attribute Support Network (CASNet), which realizes instance segmentation by predicting and clustering common attributes. 
CASNet is designed in the manner of fully convolutional and can implement training and inference from end to end. 
And CASNet manages predicting the instance without overlaps and holes, which problem exists in most of current instance segmentation algorithms.
Furthermore, it can be easily extended to panoptic segmentation through minor modifications with little computation overhead. 
CASNet builds a bridge between semantic and instance segmentation from finding pixel class ID to obtaining class and instance ID by operations on common attribute.
Through experiment for instance and panoptic segmentation, CASNet gets mAP 32.8\% and PQ 59.0\% on Cityscapes validation dataset by joint training, and mAP 36.3\% and PQ 66.1\% by separated training mode.
For panoptic segmentation, CASNet gets state-of-the-art performance on the Cityscapes validation dataset.

\end{abstract}

\IEEEpeerreviewmaketitle

\section{Introduction}
\label{sec:intro}
Image segmentation is an active computer vision research domain and the common tasks are semantic segmentation, instance segmentation, panoptic segmentation and so on. 
In Comparison with bounding box based object detection, instance segmentation can detect object at a finer pixel level granular. In contrast to semantic segmentation, instance segmentation can not  only identify class IDs but also the instance IDs.

Furthermore, panoptic segmentation can simultaneously implement instance segmentation with thing classes, such as cars, people, and with stuff classes without instances, such as sky, road, and buildings.

To date, most instance segmentation methods are derived from the two-stage bounding box detection framework, such as Faster-RCNN \cite{ren2015faster}. 
Typical two-stage detection methods first generate excess anchors, then crop the corresponding features for classification. 
Mask-RCNN \cite{he2017mask} is one of the typical method derived from two-stage framework, which gains benefit of high accuracy of bounding box detection. 
Whereas, two-stage methods usually bear the disadvantage of high computation consumption, large memory occupation, and hard train convergence problem. 

\begin{figure}[!t]
\centering

\subfloat[]{\includegraphics[width=2.5in]{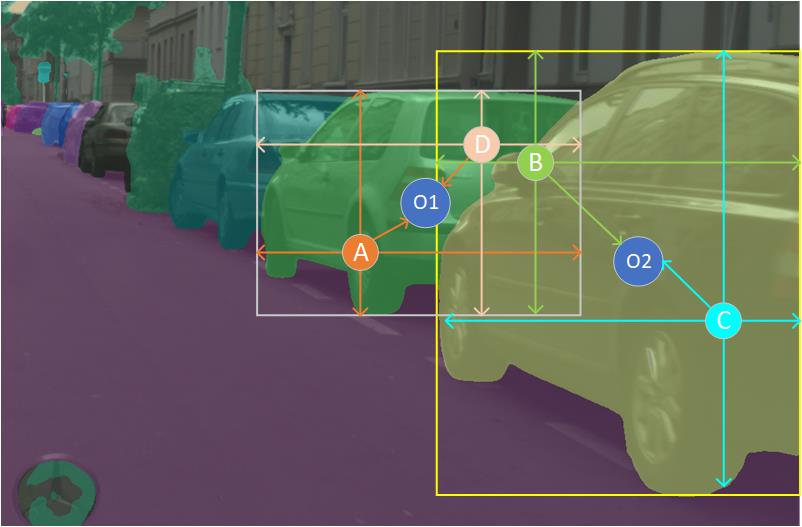}%
\label{fig_first_case}}
\quad
\centering
\subfloat[]{\includegraphics[width=2.5in]{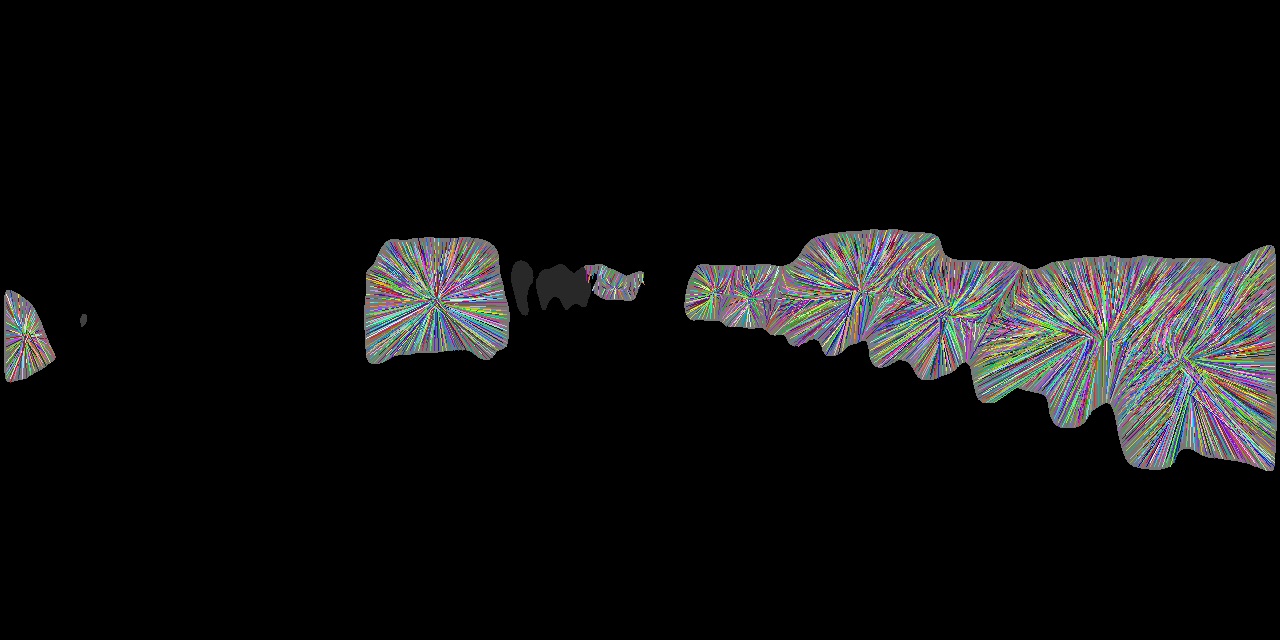}%
\label{fig_second_case}}

\caption{Illustration of the intrinsic of Mass Support Network. 
(a) Common attributes of center and distances to the bounding box. O1 and O2 are two bounding boxes' geometric centers. Points A and D both locate in the same bounding box with center O1 with the same center and bounding box O1. And points B and D both locate in the bounding box with center O2.
(b) The bounding box center predicted by CASNet.}
\label{fig:fig1}
\end{figure}

From the intuition that each pixel of an instance contains information in common of the instance, which we name it as common attribute.  
General objects in one image have common attributes, such as geometry center, weight center, the same bounding box, etc. Similarly, the common attribute of an instance can also be manually defined.
For example, as shown in Fig. \ref{fig:fig1} (a), the pixels belonging to the same car instance share the same bounding box, and geometry center.
If we can get the pixel subordination and the final instance segmentation by operating on one or more common attributes, the common attribute can be treated as a main hidden clue. 
In this paper, we propose a novel instance segmentation method by operating on these common attributes, which is debbed as Common Attribute Support Network (CASNet).
CASNet is an one-stage instance segmentation framework and works in fully convolutional way and can be trained from end to end.
Compare to RPN based methods, CASNet can realize the instance segmentation by the support of the common attribute belonging to the same instance without feature cropping and training another classifier. 
With minor modification, it can be extended to panoptic segmentation.

In contrast to other instance segmentation methods, CASNet has these advantages: 
First, CASNet is an one-stage instance segmentation framework and works in fully convolutional way, easy for train and inference. 
Next, CASNet segments instances relying on operation on common attributes. From this point of view, pixel's subordination is specific without contradiction. 
Between/among instances there are no overlaps and holes which is a common problem in present instance segmentation algorithms.
At last, during inference, many of the detected instances which fail to have any support from pixels can be removed directly and can be calibrated by the final actual pixel range.
This kind of false-negative samples are a general problem of score-based box detection methods, such as Faster RCNN, SSD, YOLO, Mask-RCNN, and so on. 

The paper has the following contributions: 
1. We propose CASNet, a one-stage instance segmentation network which works in fully convolutional way. 
2. CASNet generate comparable instance segmnetaiton results without overlaps and holes between/among instances through predicting the common attribute. 
3. With minor modification and little computation overhead, CASNet can realize high-quality panoptic segmentation. 


\section{Related work}

In recent years, instance segmentation is an active research computer vision task and deep learning-based methods have radically improved the prediction accuracy. In this section, we will introduce image instance segmentation and paoptic segmentation related works from several different perspectives.

\textbf{CRF and RNN based methods.}

Paper \cite{zhang2015instancelevel} and \cite{zhang2015monocular} ﬁrst use CNNs to perform local instance disambiguation and labeling, which are followed by a global CRF (conditional random ﬁeld) to achieve instance label consistency. Recent work \cite{arnab2016bottom} uses object detection proposals along with a deep high order CRF to classify pixels in overlapping object proposal boxes.

Since RNN(recurrent neural network) can memorize the last information, this character is utilized by researchers to realize instance segmentation, such as Romera-Paredes\cite{romera2016recurrent}.  Ren \cite{ren2017end} proposes an end-to-end RNN architecture with attention to model human-like counting process. Castrejon\cite{castrejon2017annotating} and Acuna \cite{acuna2018efficient} proposes ploygen RCNN and polygen RCNN++ instance segmentation methods based on RNN. The ConvLSTM based recurrent structures \cite{xingjian2015convolutionalLSTM} keep track of instances generated, and inhibit these regions from further instance generation. Additionally, \cite{uhrig2016pixellevel} extracts image features and employs a pipeline including a ConvLSTM structure to direct a bounding box generation network followed by individual instances extracting segmentation network.

\textbf{Instance embedding approaches.} 
Inspired by heuristic clustering, instance embedding-based methods are proposed, such as  \cite{de2017semantic}, \cite{fathi2017semantic}, \cite{kong2018recurrent}. These approaches map each pixel of the input image into an embedding space by the neural network. The network learns to map the same instance's pixels into nearby points in the embedding space, while pixels of different objects are mapped into distant points in the embedding space. However, currently, this kind of method demonstrates inferior performance on standard benchmarks compared to detection-first methods.

\textbf{Detection-first instance segmentation methods.} 
Recently, instance segmentation approaches based on standard object detection pipeline have become the mainstream, achieving state-of-the-art results on various benchmarks. Dai proposes \cite{dai2015instanceaware} a feature masking method, where a multi-task network cascade (MNC) is used to implement instance segmentation. \cite{dai2016instancesensitive} proposes a position-sensitive method which is similar to R-FCN \cite{dai2016rfcn} in a fully convolutional way. A more developed method \cite{li2016fcis} works also in this way to get the instance results from the combination of every position-sensitive part. MaskLab \cite{chen2018masklab} is another approach of the position-sensitive method to segment instances. Mask-RCNN \cite{he2017mask} is one of the typical object detection based instance segmentation methods, which fully uses the effectiveness of the region proposal network widely used on Faster-RCNN \cite{ren2015faster}. Recently, many variation of this pipeline are brought up based on , such as \cite{liu2018path_panet}, \cite{chen2019hybrid}, \cite{hariharan2016object}.

\textbf{Other instance segmentation methods.} 
Paper \cite{bai2017watershed} uses watershed method to segment instances in an image. \cite{chen2017dcan} uses contour-aware networks, \cite{hayder2017boundary} uses boundary-aware method, \cite{ye2017depth} uses depth-aware method, and \cite{liu2017sgn} uses sequential grouping method to segment instances. Box2Pix \cite{uhrig2018box2pix} is a one-stage method to predict the bounding boxes and centers of each pixel. AdaptIS \cite{sofiiuk2019adaptis} adopts instance batch normalization which is widely used in style transfer to segment instances. Inspired by YOLO, \cite{bolya2019yolact} proposes an one-stage instance segmentation method. SOLO \cite{wang2019solo} is another position-sensitive method which adopts instance batch normalization. Additionally, 3d points cloud instance segmentation and RGB-D data methods are proposed, such as \cite{wang2018sgpn}, \cite{hou20193d}, \cite{yi2019gspn}, and \cite{gupta2014learning}.

\textbf{Panoptic segmentation.}
Panoptic segmentation is brought up by Kirillov etc. \cite{kirillov2019panoptic}, and the evaluation metric is also depicted in detail. Paper \cite{de2018panoptic} simply joins semantic segmentation and instance segmentation together to realize the panoptic segmentation. Li etc. proposes TASCNet \cite{li2018learning} to realize this task and a weakly supervised network using CRF and PSPNet is brought up in \cite{li2018weakly}. UPSNet \cite{xiong2019upsnet} is a more compact method to realize panoptic segmentation. Recently, one-stage work such as AdatIS \cite{sofiiuk2019adaptis} can directly infers all the thing and stuff classes without semantic or instance segmentation at first.

\section{Method}

\begin{figure*}
\begin{center}
\includegraphics[width=14.0cm]{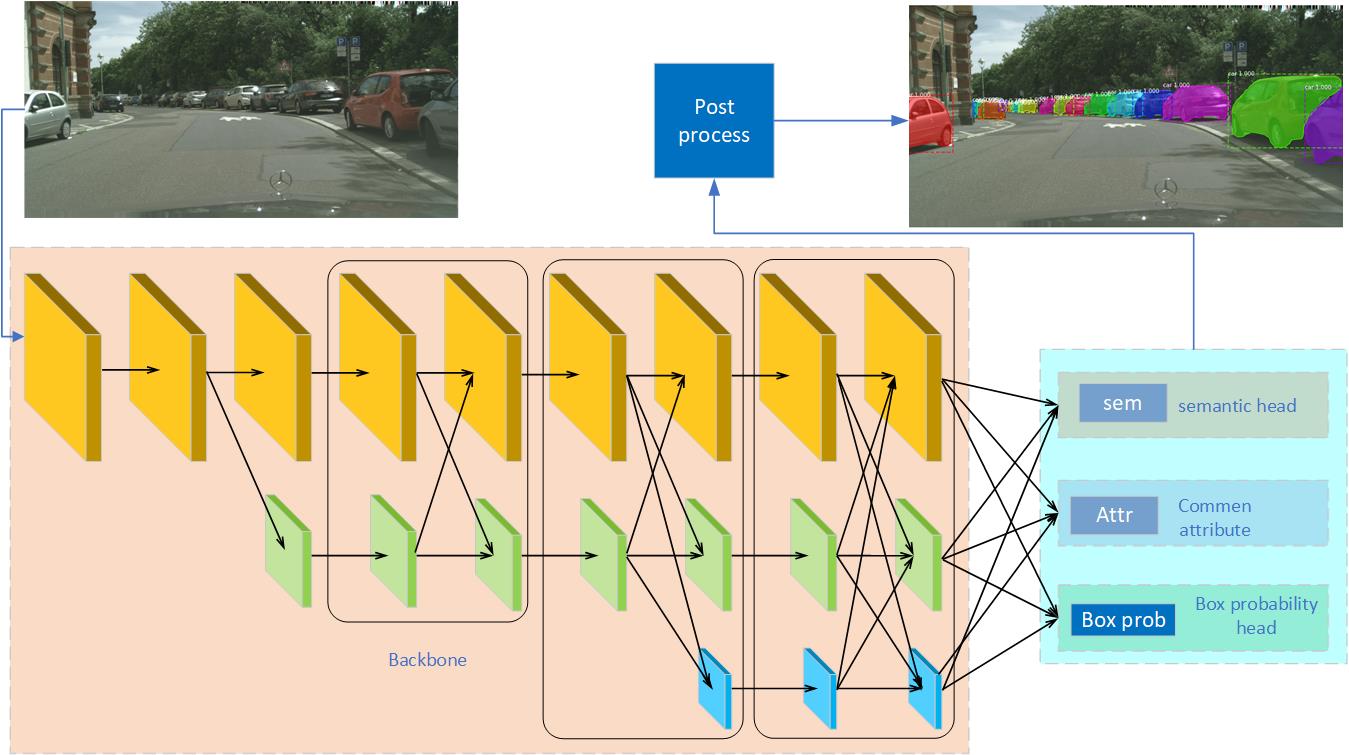}

\end{center}
\caption{Structure of Common Attribute Support Network.}
\label{fig:fig2}
\end{figure*}

In instance and panoptic segmentation, classes consist of thing and stuff, where things (objects with a well-defined shape, e.g. car, person) and stuff (amorphous background regions, e.g. grass, sky).
Let \(x \in {\chi} = R^{H \times W \times 3}\) be an image and \(u \in \omega = \{1,...,H\} \times \{1,...,W\}\) a pixel. In instance and panoptic segmentation, the goal is to map the image to a collection \(S_{N} = \{S_1,...,S_k, S_{k+1}, S_{N}\} \subset 2^{\omega}\) of image regions,  each in \({S_1,...S_k}\) representing an occurrence of a 'thing' object of interest and each element in \({S_{k+1},...S_{N}}\) representing the 'stuff' groups. 
In instance segmentation, all the stuff classes will be treated as one background region that is denoted as \(S_{0} = \Omega-\cup_k S_k\).  The regions as well as their number are a function of the image and the goal is to predict both.

Inspired by the property of pixels belong to one instance has common attributes, e.g. geometry center, bounding box and so on, we propose a novel algorithm CASNet that differs instances by clustering pixels by their common attributes. First, the thing and stuff class labels without instance ID can be got by one step semantic segmentation. Then, the clustering algorithm labels each pixel an instance ID by the predicted common attributes. 

The principle of CASNet is illustrated in Fig. \ref{fig:fig1} (a). Points of A and D belong to instance O1, B and D belong to instance O2. During semantic segmentation, all the points of A, B, C, and D are predicted as one class as 'car'. The clustering process of CASNet can implement labeling pixels an instance ID by the support of each pixel's common attributes. For example, B and D are lying in an area intersected by two bounding boxes corresponding to instance O1 and O2, but they are labeled two instance IDs since they predict different common attributes for the instance they belong to.

\subsection{Network structure}

The structure of CASNet is demonstrated in Fig. \ref{fig:fig2}, it consists of backbone, semantic head, common attribute head, and box probability head. To acquire the final result, a post process step is needed to fully use the predicted common attributes.

Any mainstream CNNs, such as ResNet, UNet, MobileNet, can be used in CASNet.
Since CASNet uses semantic segmentation as the first process, backbones adopted for semantic segmentation are more fit for this network. 
To achieve a better semantic segmentation result, HRNet \cite{sun2019deep} is choosen as the backbone. 
The HRNet maintains high-resolution representations by connecting high-to-low resolution convolutions in parallel and strengthens high-resolution representations by repeatedly performing multi-scale fusions across parallel convolutions. 
In our configuration, HRNet, whose structure can refer to \ref{fig:fig2}, outputs a feature pyramid which is composed of three sizes feature maps of 1/2, 1/4, 1/8 the size of input.

The semantic head is configured the same as FCN. 
The fully convolutional layer outputs the number of class logits according to the classes for recognition. 
If only implementing instance segmentation, only the thing classes are needed to segment, all the stuff classes are treated as background, or under the scenario of panoptic segmentation, this head outputs class numbers the same as semantic segmentation. 
This is the minor difference of realization between instance segmentation and panoptic segmentation, which only adds little computation cost.

The common attribute head, the same as semantic segmentation head, also works in fully convolutional way, and it outputs common attributes of each instance. 
Here we choose the four distances to the four borderlines of bounding box corresponding to the instance as common attributes. 
We also consider the geometry center as a common attribute. 
Since the center can be calculated by the borderline distances, the center does not need extra output in training. 
The FCOS \cite{tian2019fcos} also predicts the four borderline distances to realize the bounding box detection. 
Difference between FCOS and CASNet is that we use all the points' common attributes as a criterion to realize instance segmentation, while FCOS only use the center points' common attributes to realize bounding box prediction.

To acquire the seed boxes, we adopt the box probability head which is also used in FCOS \cite{tian2019fcos}, which is designed to predict the box center probability of each position. 
It also works in the fully convolutional and class-specific way. 
All the CASNet components in the process of training is the backbone, semantic head, common attribute head, and box probability head. 
To obtain the final result during inference, a post process procedure is also needed which can refer to Section Post processing.

\subsection{Losses}
During the training of CASNet, three losses are defined for semantic segmentation, common attribute head, and box probability head. For semantic segmentation head, CE (Cross-Entropy) loss is adopted, which is widely used in semantic segmentation tasks. 
When only the instance segmentation is implemented, the ground truth contains the thing classes with the stuff classes are all treated as background. Under the panoptic segmentation, the ground truth assigns label IDs as the semantic segmentation does. The CE loss can refer to Equ. \ref{eq1}.
\begin{align}\label{eq1}
L_{cls} = \ell(x, class) = -\log\left(\frac{\exp(x[class])}{\sum_j \exp(x[j])}\right)
\end{align}
where x[class] means the probability of the predicted logits for class. 

The common attribute head predicts the four distances to its corresponding bounding box's four borderlines. The ground truth is divided by a fixed number according to the input size, e.g. the number is 8 when the input size is 1/8 of input image size. The L1 Loss is adopted as  shown in Equ. \ref{eq2}. 

\begin{equation}
\begin{aligned}\label{eq2}
L_{common} = {mean}(\ell(x, y))  = {mean}(\left[l_1,\dots,l_N\right]^\top),\\
l_n = \left| x_n - y_n \right|
\end{aligned}
\end{equation}

The box probability head outputs the probability for each pixel and the ground truth is defined as that only the small area in the center is set as probability 1, otherwise as 0. Then only the center of box will output the high probability. This strategy is proven to be effective in FCOS. The BCE (binary cross-entropy) loss as in Equ. \ref{eq3} is used for box probability head. 

\begin{align}\label{eq3}
        L_{prob} = - w_n \left[ y_n \cdot \log \sigma(x_n)
        + (1 - y_n) \cdot \log (1 - \sigma(x_n)) \right],
\end{align}

The total loss is a weighted sum of the three losses shown in Equ. \ref{eq4}.

\begin{align}\label{eq4}
L_{total} = \alpha L_{cls} + \beta L_{common} + \gamma L_{prob}
\end{align}
where ${\alpha}$,  ${\beta}$, and ${\gamma \in}$ (0,1).

\subsection{Post processing}
During inference, the semantic segmentation result can be get from an argmax operation.
Next, the instance will be calculated in class specific way. 
Take class 'car' as an example, the pixels predicted as car marked as \(S_{car}\). Each pixel in \(S_{car}\) will calculate its bounding box \(B_{car}\) and geometry center \(C_{car}\) from the common attribute head prediction of four borderline distances. 
Then, from the box probability head, CASNet chooses points above a threshold, e.g. 0.5, and operates by the NMS (non-max suppression) operation. The outputs are the seed points \(P_{seed}\) and its corresponding boxes \(B_{seed\_car}\). 
Intersection of Union (IoU) ratio will be calculated between \(B_{car}\) and \(B_{seed\_car}\). Each pixel in \(S_{car}\) will give its vote in \(P_{seed}\) by the IoU ratio, the most prominent one will be its final classification. Assume \(k_{car}\) elements in \(P_{seed}\), the instance output will be \({S_1, S_2, ..., S_{k_{car}}}\). 
The vote mechanism is called common attribute support.
For the center prediction, once two IoU ratio is approximate the same, center will give an direction for the classification. Fig. {\ref{fig:fig1} (b) gives a demonstration of center prediction}.

Finally, all the pixels in \(S_{car}\) has its unique instance ID, this is why CASNet can generate instance results without overlapped area and holes. In contrast, the overlapped area and holes between/among instances is an ubiquitous problem for most of present instance segmentation algorithms. 

Noticeable, many of the seed boxes without any support from pixels can be removed directly. These false-negative boxes are a common problem for score-based box detection methods, such as Faster-RCNN, SSD, YOLO, Mask-RCNN, and so on. This is another advantage of CASNet.


\section{Experiments}
Targeting for Autonomous Driving and ADAS, CASNet is tested on the Cityscapes dataset, which focuses on traffic environment understanding. This dataset has 5,000 fine-annotated images of ego-centric driving scenarios in urban settings that are split into 2,975, 500, and 1,525 images for training, validation, and testing split respectively. Cityscapes consists of 8 and 11 classes for thing and stuff, respectively. Experiments of CASNet is trained on the train split set and evaluated on the validation set. All images in Cityscapes have the same resolution 1024\(\times\)2048.

\textbf{Implementation details.} 
Implementation of CASNet is based on PyTorch with 8 GPUs of 2080Ti. The batch size is 8 and the input image resizing and cropping operation are adopted for memory saving. The short side of the input image is resized from 512 to 1024 and the crop size is 640\(\times\)1280.
CASNet loads pre-trained weight for HRNet and trains for 24 epochs and drops the learning rate by a factor of 10 at 18th and 22nd epochs.  
We use loss accumulation strategy and SGD with the initial learning rate 0.001 for pre-trained backbone from HRNet and all other parameters.
The loss weights are  set to ${\alpha}$=1.0, ${\beta}$=1.0, and ${\gamma}$=1.0 under our experiments.
To make things more clear, two modes are adopted during inference. The first one is joint training that all three heads are trained jointly with the backbone. The second one is separated training, which semantic results are obtained from the prediction of a normal HRNet trained on Cityscapes.

\textbf{Instance segmentation result.} 

For Cityscapes, we report instance and panoptic results on the validation and test set.
To evaluate the instance segmentation performance, we adopt mean Average Precision (mAP) as defined by Cityscapes \footnote{https://www.cityscapes-dataset.com/benchmarks/instance-level-results}. Through minor modification with the semantic head to segment all the classes thing and stuff (19 classes), we can easily get the panoptic segmentation. 
The evaluation metric for panoptic segmentation is panoptic quality (PQ), recognition quality (RQ), and semantic quality (SQ), which can refer to \cite{kirillov2019panoptic}.

The instance experiment results and comparison with other main algorithms are demonstrated in Tab. \ref{tab:tab_ins}.
 The mAP (Val) of CASNet with joint training is ${32.8\%}$ which is a little better than the typical two-stage detector Mask-RCNN (ResNet 50, without COCO pretrained) 31.5\% \cite{liu2018path_panet}. Since CASNet adopts the HRNet as backbone while Mask-RCNN adopts ResNet, the comparison is just a qualitative illustration of the effective of algorithm.
AdaptIS \cite{sofiiuk2019adaptis} is an one-stage detector which creatively adopts instance normalization that widely used for style transfer for instance segmentation. 
The result of CASNet  with joint training is ${32.8\%}$ that is \(0.5\%\) higher than AdaptIS ${32.3\%}$. 
Experiment shows that the joint training deteriorate the semantic segmentation accuracy. When implementing separated training, the semantic segmentation results obtained from the HRNet that sololy servers as semantic segmentation. Results demonstrates \(3.5\%\) improvement than joint training. This result remind us that the common attribute prediction and the semantic segmentation has a better way of joining together without precision decline. More exploration to boost the precision of segmentation could be explored.

\begin{table}
\caption{Instance segmentation result of CASNet}
\begin{center}
\begin{tabular}{|l|cc|cc|} 
\hline
Method &mask AP&$AP_{50}$&mAP(val) &$AP_{50}$(val) \\
\hline\hline
SGN\cite{liu2018path_panet}
      &0.250  &0.449    &0.292        &- \\
Mask R-CNN R50\cite{liu2018path_panet}
      &0.262  &0.499    &0.315        &- \\
SegNet\cite{liu2018path_panet}                
      &0.295  &0.556    &-            &- \\
PANet[fine-only]\cite{liu2018path_panet}
      &\textbf{0.318}  &\textbf{0.571}    &\textbf{0.365}        & -\\
SAIS\cite{arnab2017pixelwise}     
      &0.174  &0.367    &-            & -\\       
Pixel Encoding\cite{arnab2017pixelwise}     
      &0.089  &0.211    &-            & -\\    
pixelwise\cite{arnab2017pixelwise}     
      &0.20   &0.388     &-            &- \\   
AdaptIS(R50)\cite{sofiiuk2019adaptis}
      &-      &-        &0.323        &-\\
AdaptIS(R101)\cite{sofiiuk2019adaptis}
      &-      &-        &0.339        &-\\
UPSNet(R50)\cite{Xiong_2019}
      &-      &-        &0.339        &-\\
CASNet(joint)
      &0.2527 &0.4780   &0.328        &0.566\\ 
CASNet(separated)
      &0.261  &0.480    &0.363        &\textbf{0.599}\\

\hline
\end{tabular}
\end{center}

\label{tab:tab_ins}
\end{table}

\textbf{Ablation study of common attribute.}

\begin{figure}[!t]
\centering

\subfloat[]{\includegraphics[width=1.5in]{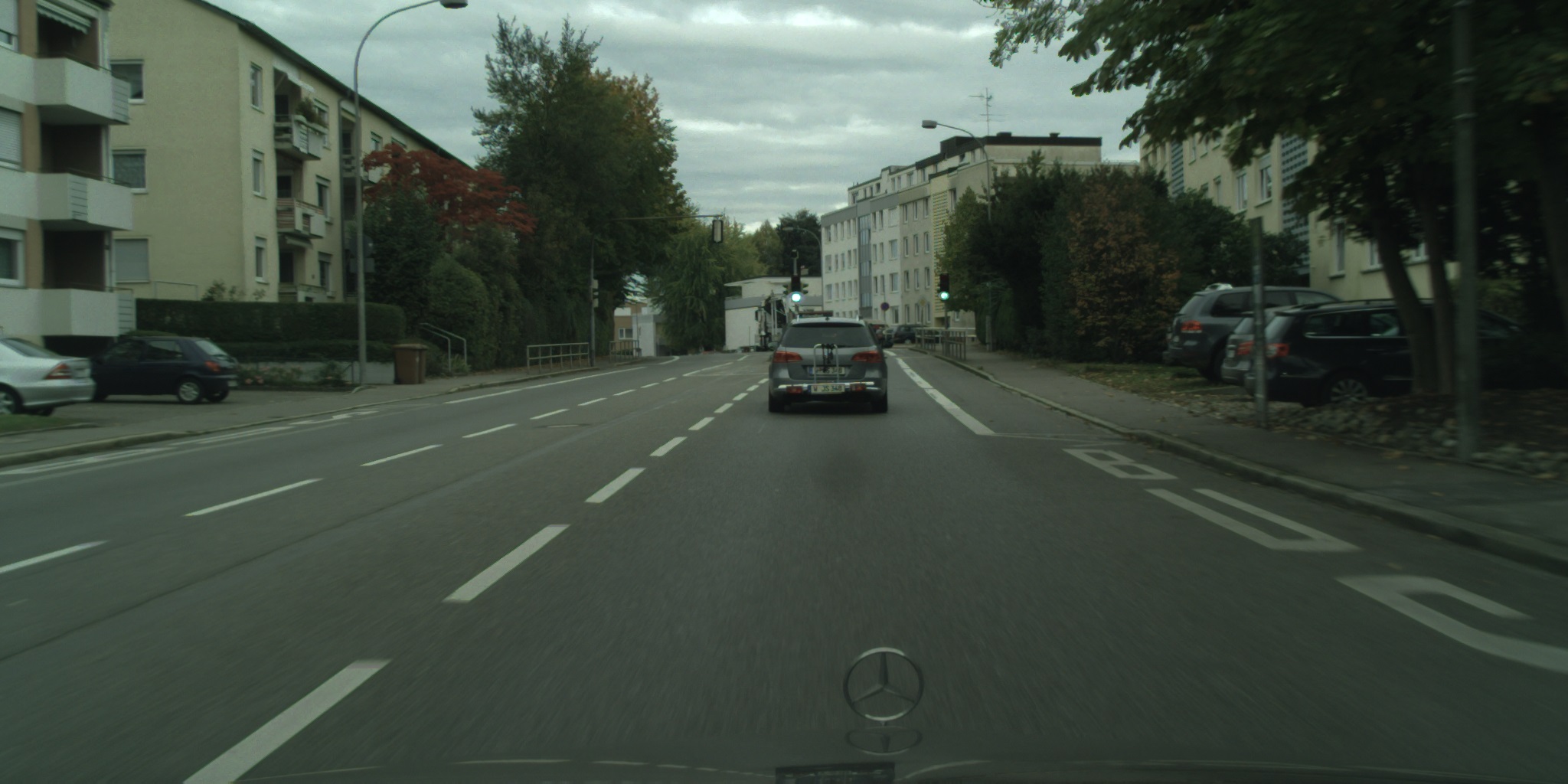}%
\label{fig_first_case}}
\centering
\subfloat[]{\includegraphics[width=1.5in]{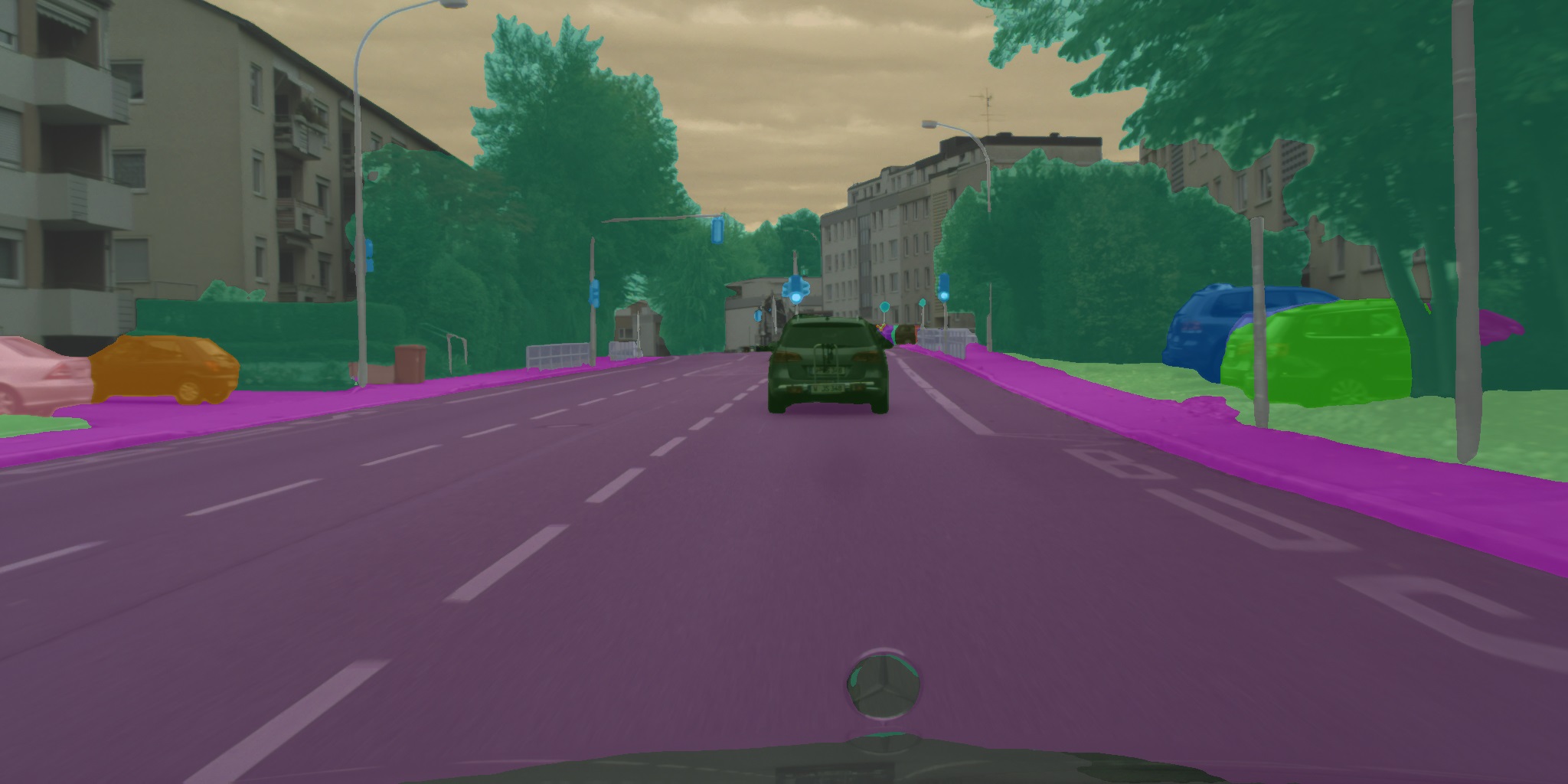}%
\label{fig_second_case}}
\quad

\centering
\subfloat[]{\includegraphics[width=1.5in]{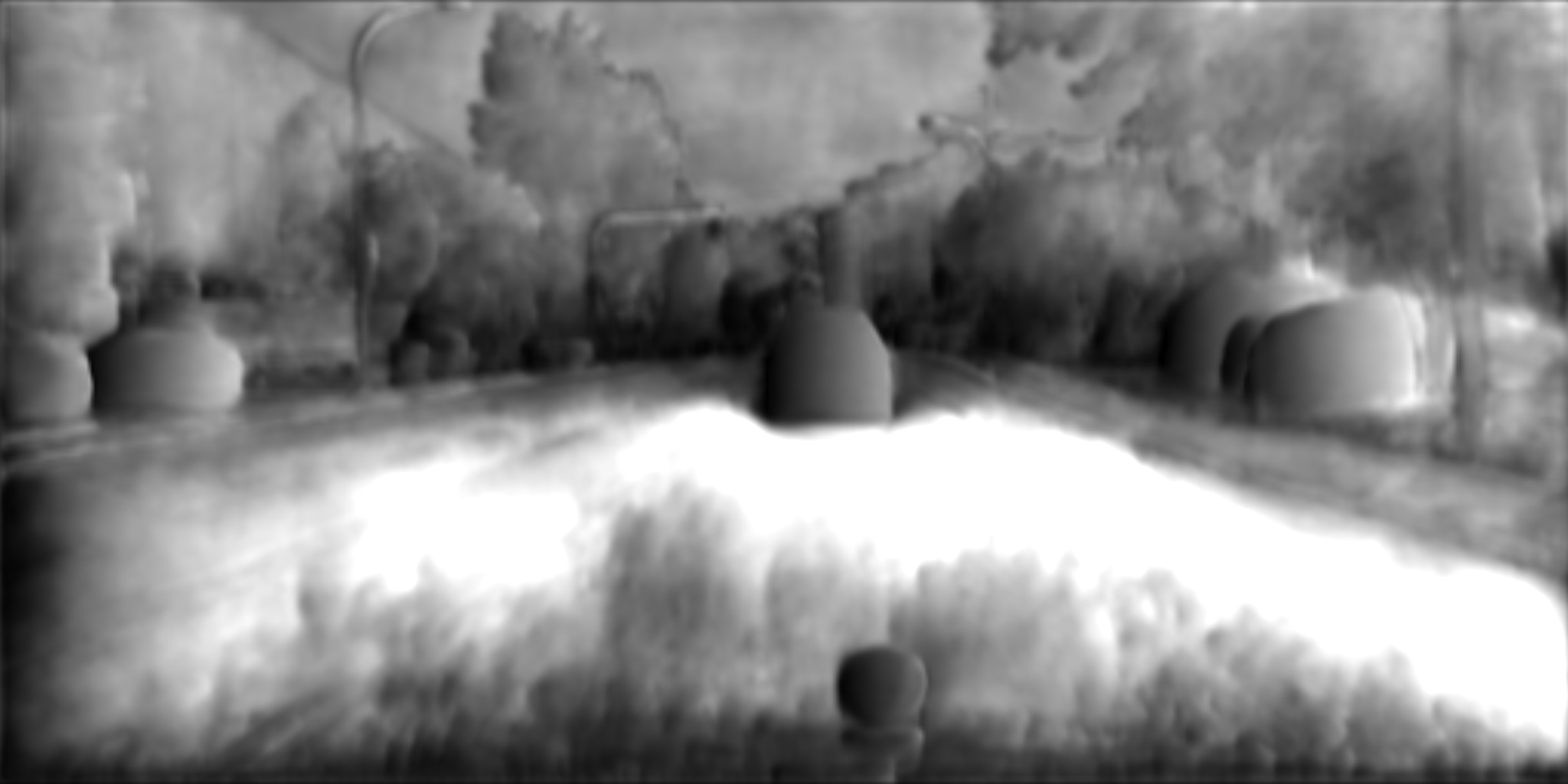}%
\label{fig_third_case}}
\subfloat[]{\includegraphics[width=1.5in]{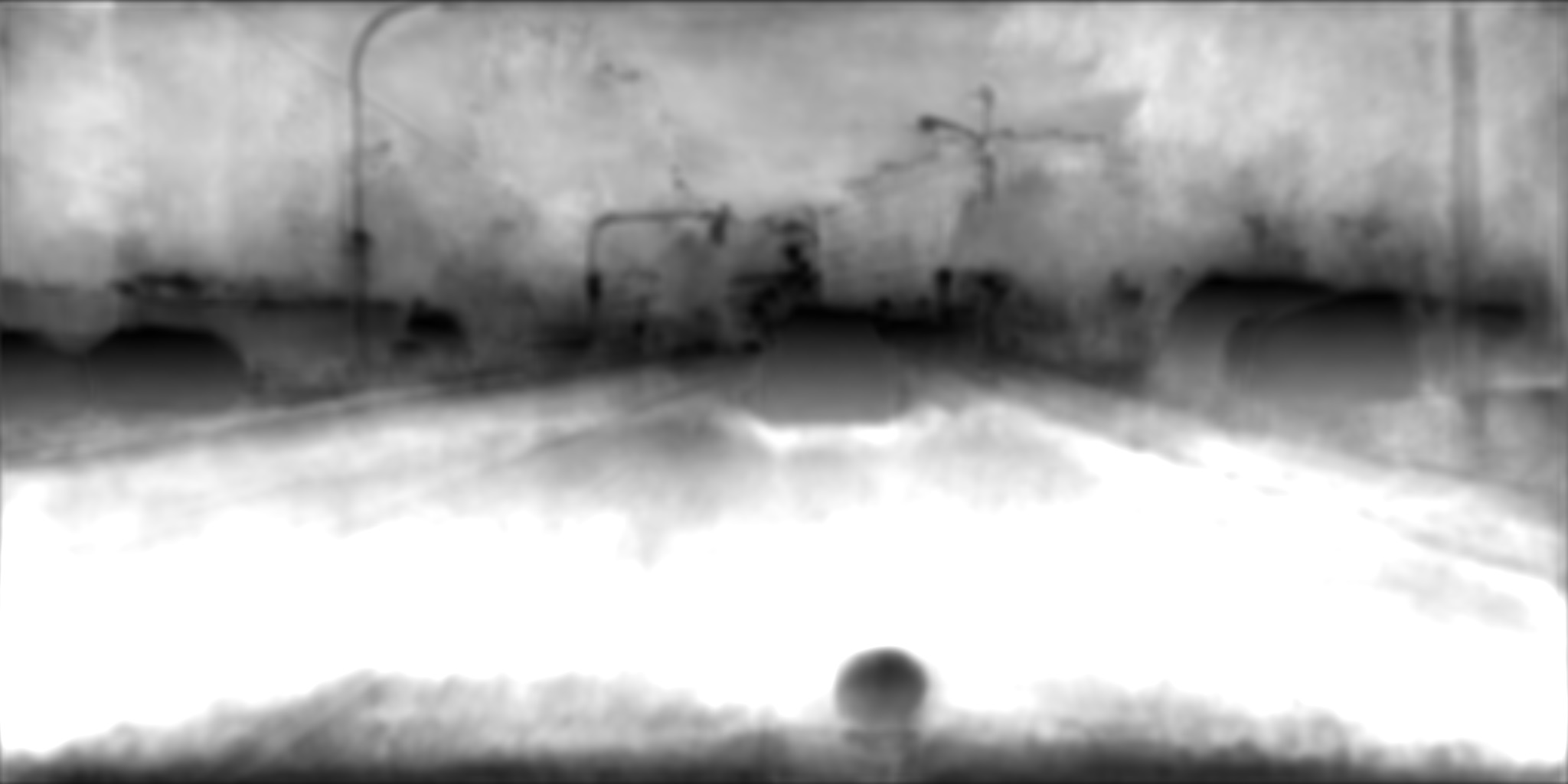}%
\label{fig_fourth_case}}
\quad

\centering
\subfloat[]{\includegraphics[width=1.5in]{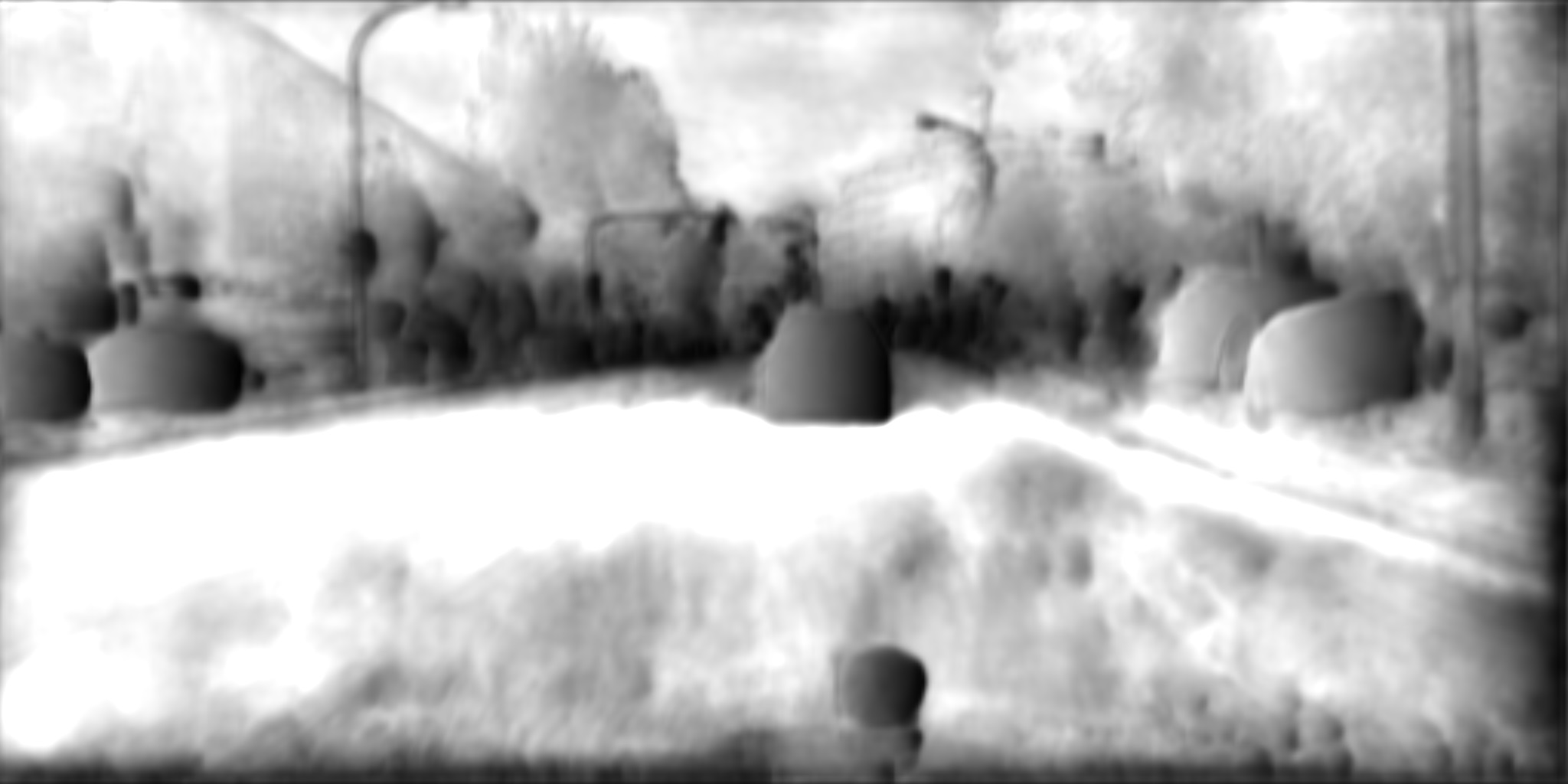}%
\label{fig_fifth_case}}
\centering
\subfloat[]{\includegraphics[width=1.5in]{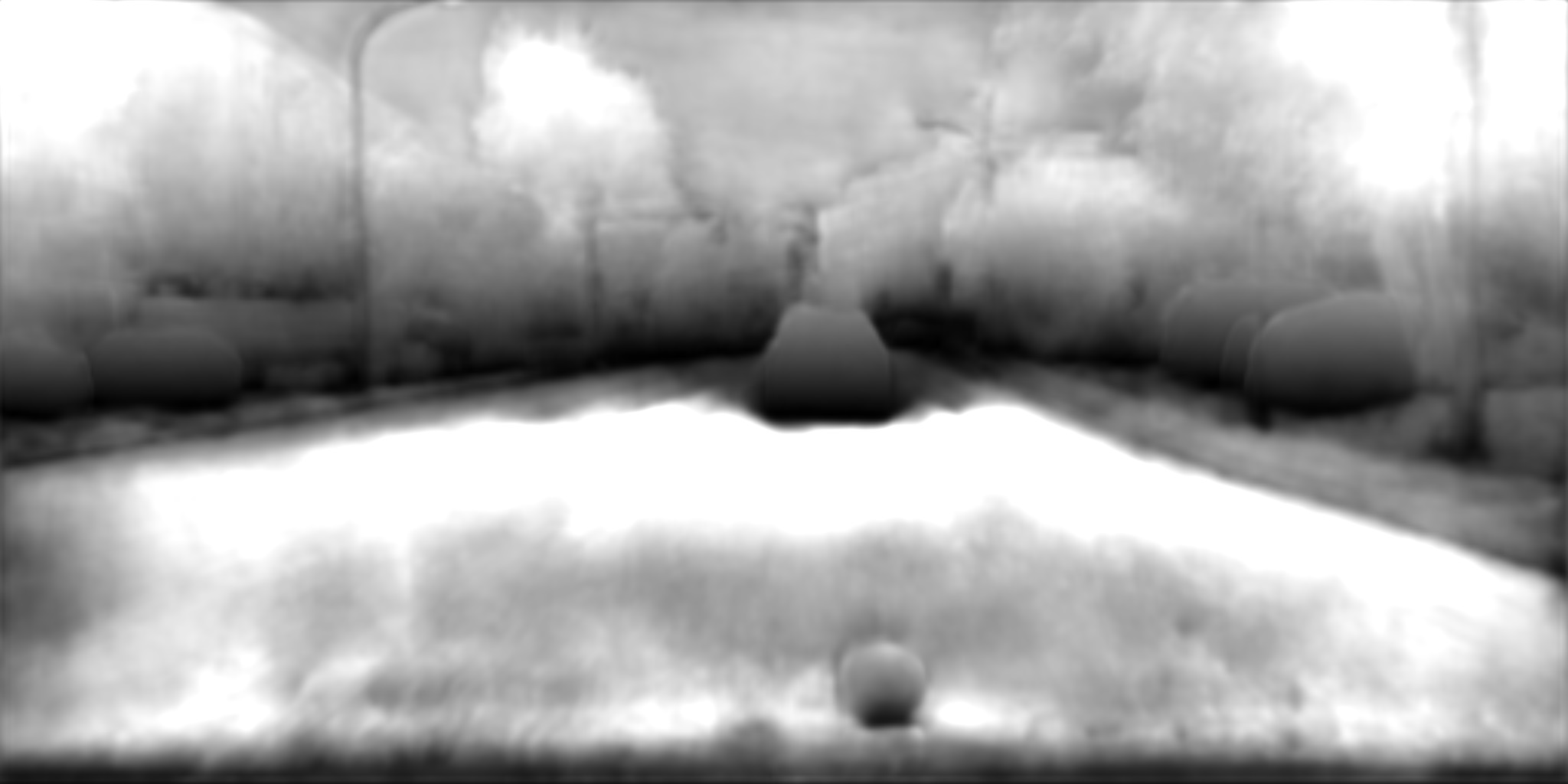}%
\label{fig_sixth_case}}
\caption{Illustration of common attribute prediction. (a) is an original image from Cityscpapes dataset. (b)(c)(f)(e) are the predictions of left, up, right, and bottom bounding box distances. Images are viewed in gray and pixel with brighter value means larger distance. (d) is the final output of the segmentation.}
\label{fig:reg}
\end{figure}

CASNet is inspired from the intuition that each instance point contains common attribute of current instance. 
FCOS \cite{tian2019fcos} also takes this assumption which only takes points around object center.
However, we go further and bolder to use all the points' common attribute information to take the more challenging instance segmentation task other than object detection. 
How well does the simple idea that the common attribute works at instance level and can the pixels at the edge of object can find its subordination instances? 
Fig. \ref{fig:fig1} (b) is a visualization of center prediction from the common attribute head. we can find that the predicted centers of objects are almost all pointing to the geometry center of the instance even the point locate at the edge area of the object. 
More results demonstrate that the common attribute prediction is effective for instance segmentation. 
The distances predicted from the common attribute head to the top, bottom, right, and left side are illustrated in Fig. \ref{fig:reg}. We can see that pixels adjacent but from different instances have a clear distance prediction which is visualized as a boundary in the image.
This is the basis for CASNet of realizing the instance segmentation by clustering the common attributes just from its pixel.

\begin{figure}[!t]
\centering

\subfloat[]{\includegraphics[width=1.5in]{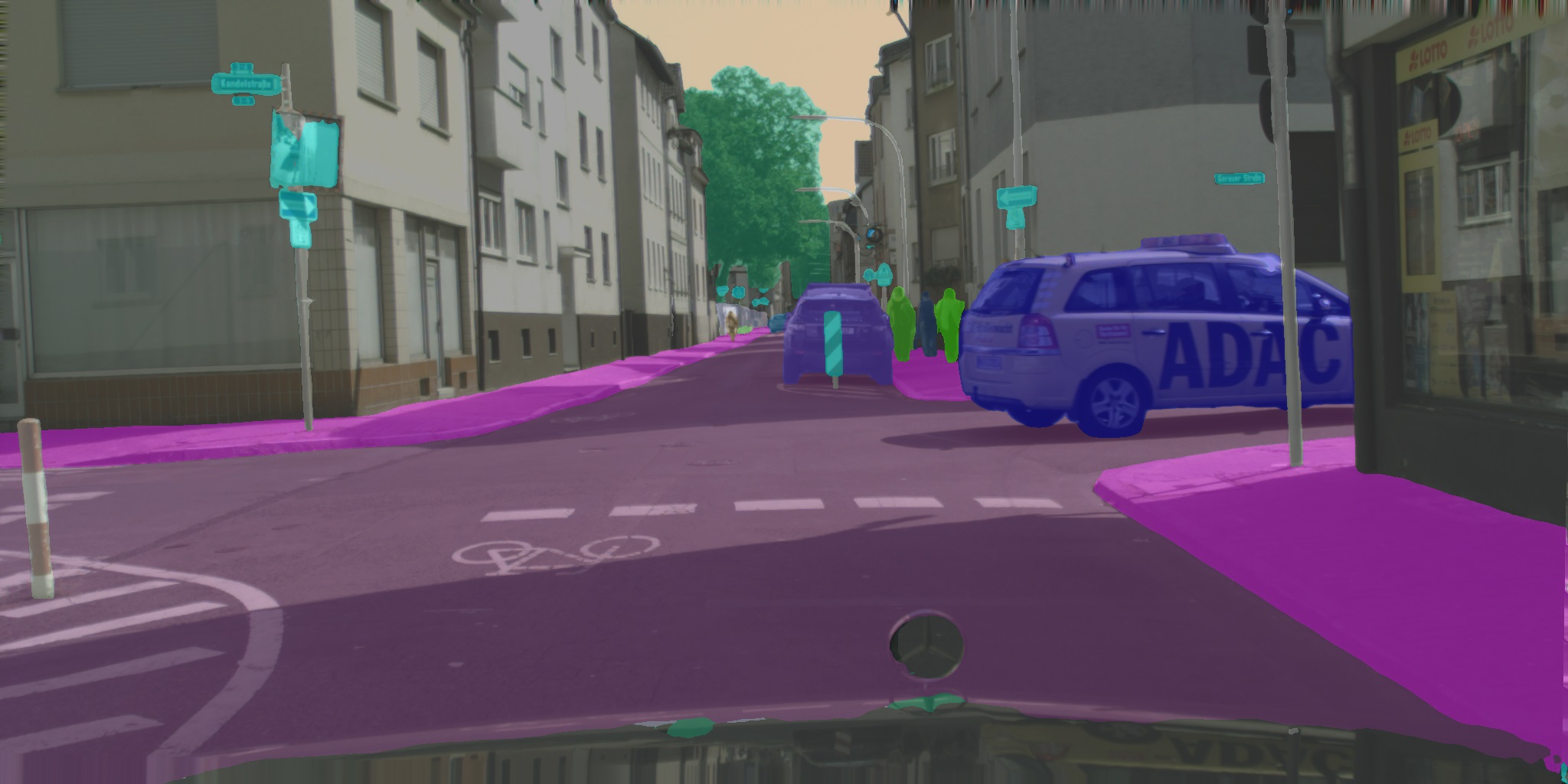}%
\label{fig_first_case}}
\centering
\subfloat[]{\includegraphics[width=1.5in]{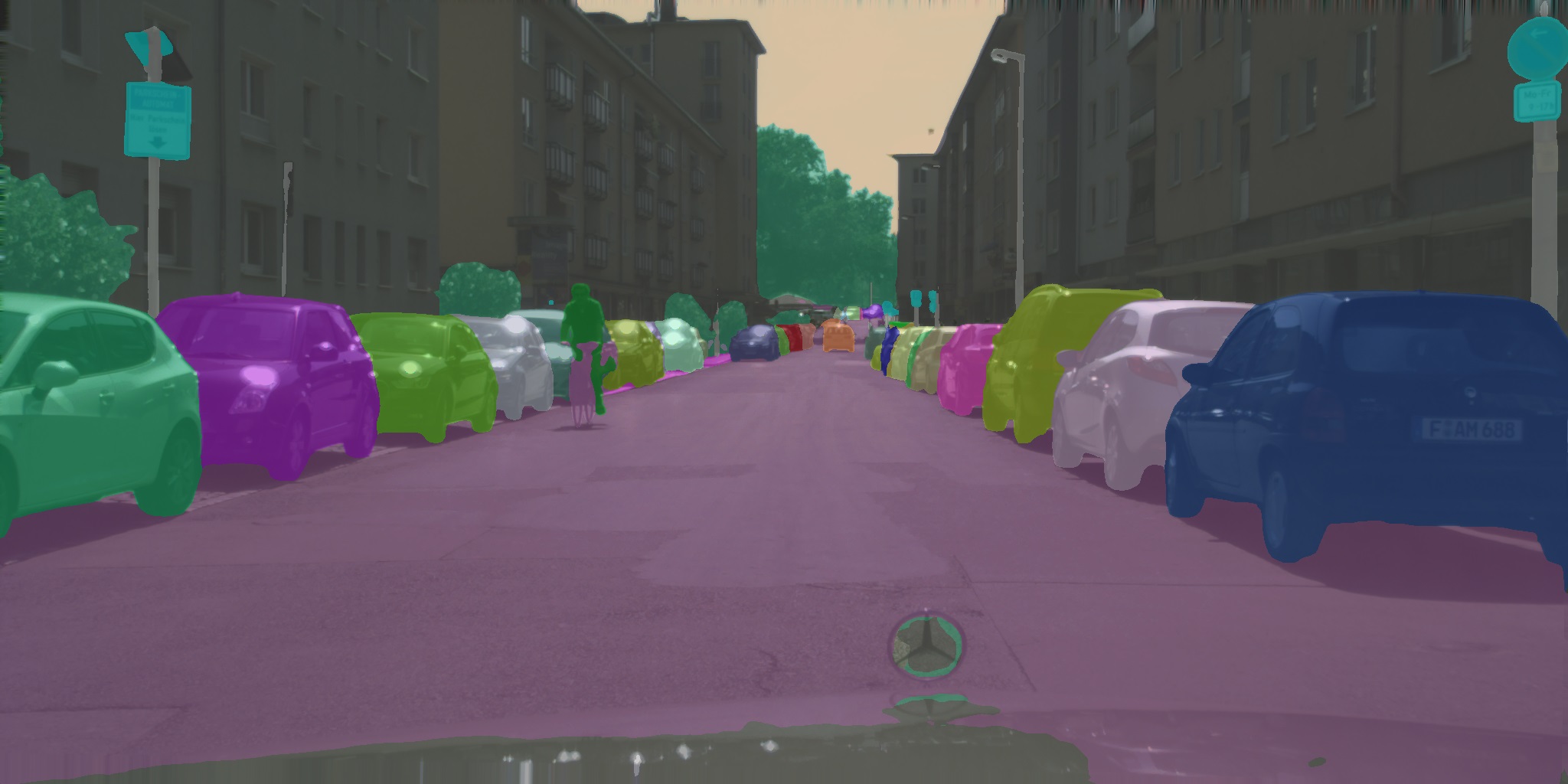}%
\label{fig_second_case}}
\quad

\centering
\subfloat[]{\includegraphics[width=1.5in]{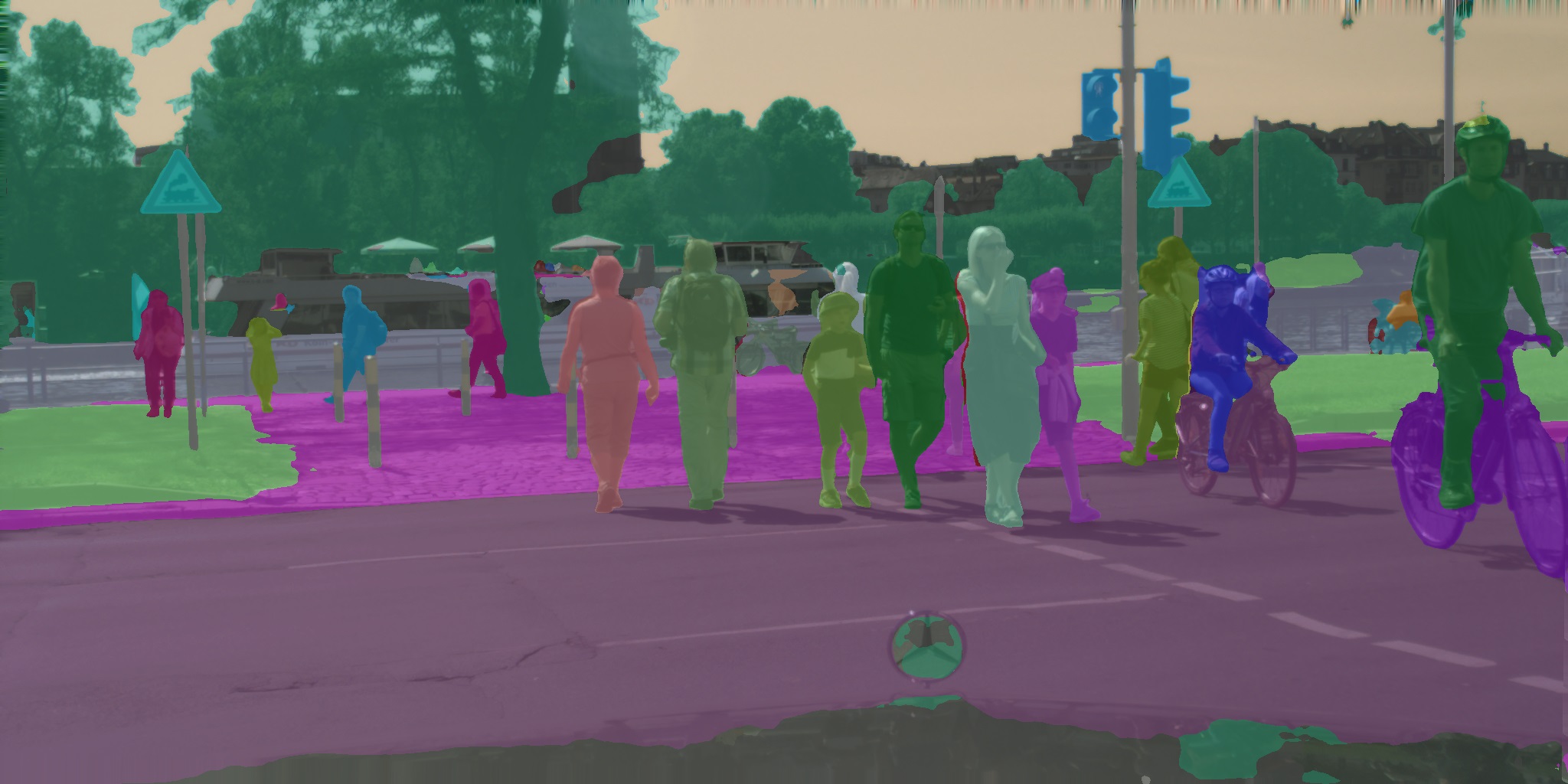}%
\label{fig_third_case}}
\subfloat[]{\includegraphics[width=1.5in]{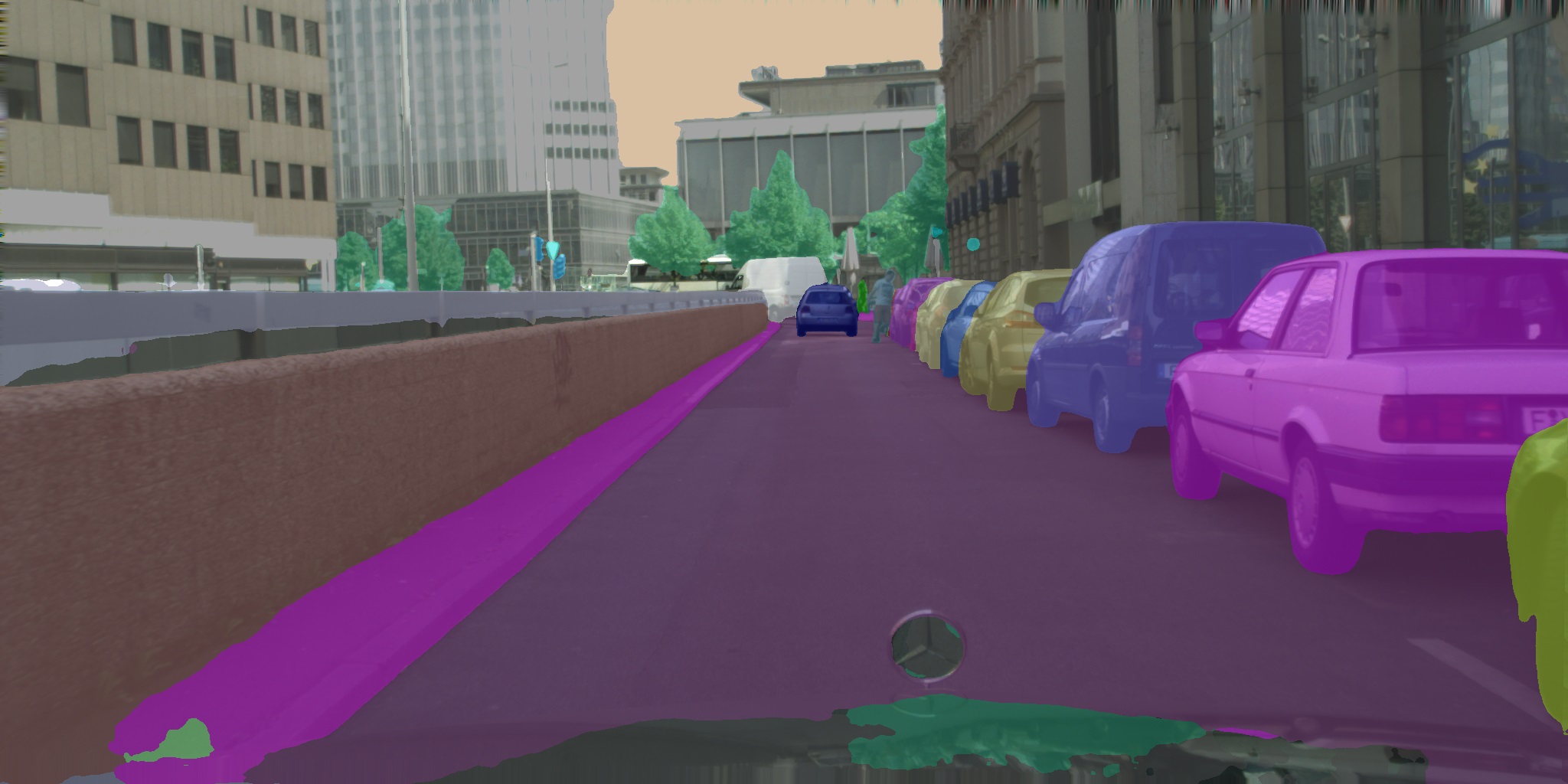}%
\label{fig_fourth_case}}
\quad

\centering
\subfloat[]{\includegraphics[width=1.5in]{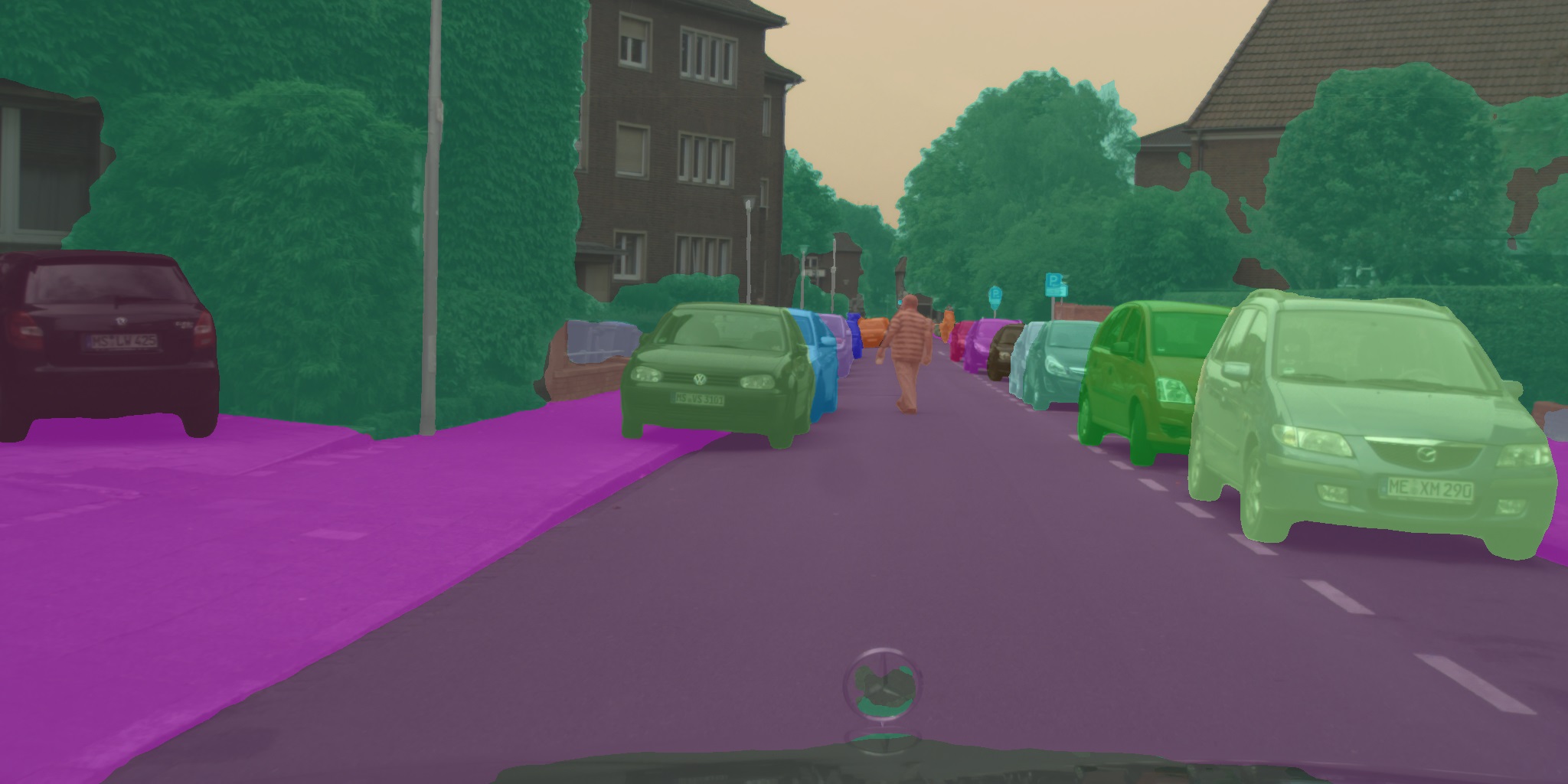}%
\label{fig_fifth_case}}
\centering
\subfloat[]{\includegraphics[width=1.5in]{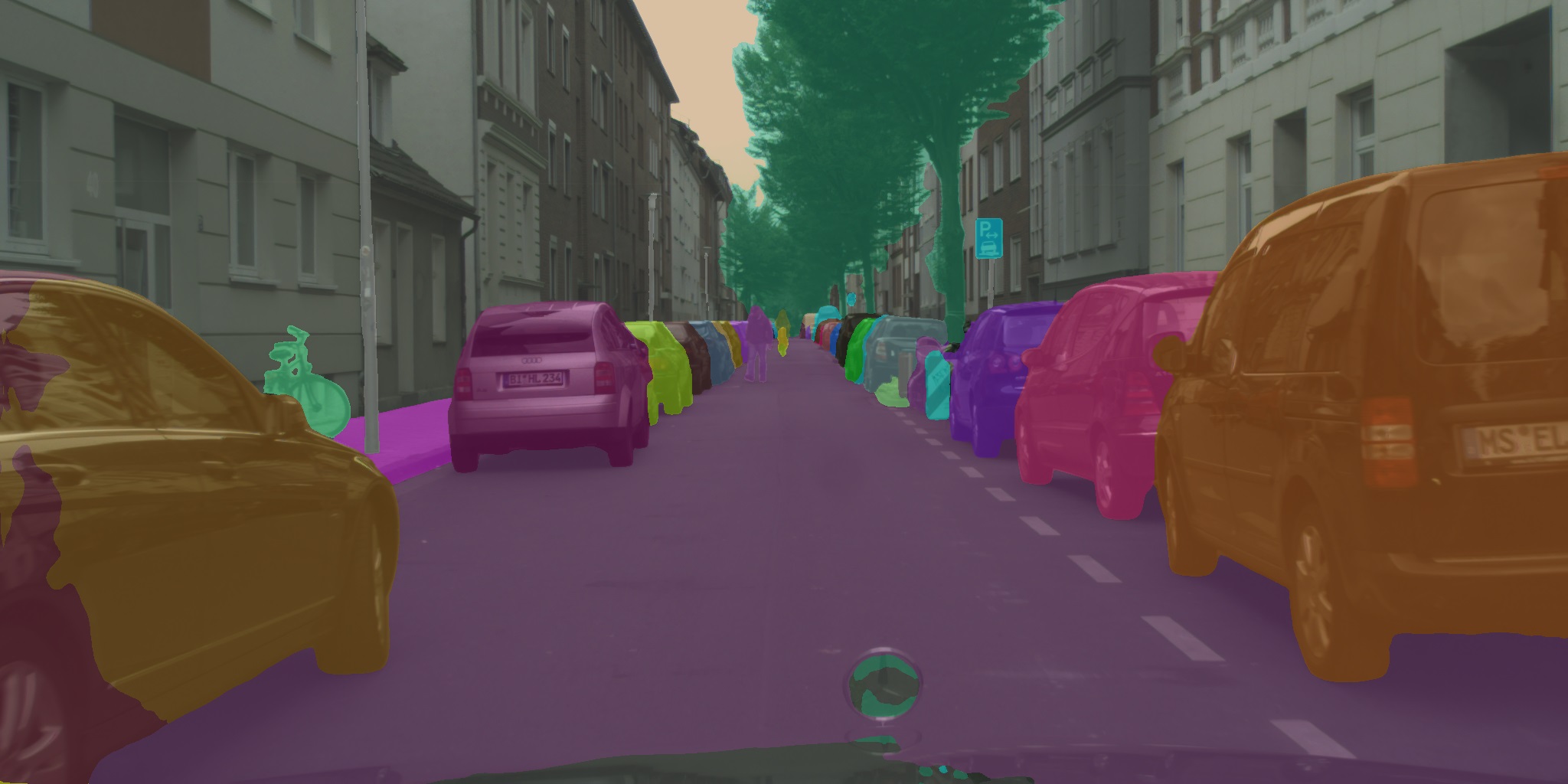}%
\label{fig_sixth_case}}
\caption{Results of panoptic segmentation.}
\label{fig:results}
\end{figure}

\textbf{Panoptic segmentation result.}
Panoptic segmentation can be realized only by adding the predicting of stuff classes at the semantic segmentation head. This will cause little computation overhead than realizing instance segmentation by CASNet. The same to instance segmentation, the panoptic segmentation is also without overlaps and holes. 
The results of panoptic segmentation are shown in Tab. \ref{tab:tab_pan}. 
We can see CASNet with separated training achieved PQ 66.1\% Cityscapes validation set which is the best performance over the methods and this is 4.7\% higher than UTIPS \cite{li2020unifying} and 4.3\% higher than UPSNet-M-COCO. 
The panoptic segmentation is benefit from the high accuracy of stuff class prediction.
More panoptic segmentation results for visualization can refer to Fig. \ref{fig:results}.

\begin{table*}
\caption{The panoptic segmentation results of CASNet}
\begin{center}
\begin{tabular}{|l|ccc|cc|cc|} 
\hline
Models &PQ   &SQ  &RQ  &$PQ^{Th}$ & $PQ^{St}$& mIoU  & AP\\
\hline\hline

CRF + PSPNet \cite{li2018weakly}
      &0.538 & -  &-   & 0.425      & 0.621     &0.716  &0.286\\
MR-CNN-PSP    
      &0.580 &0.792&0.718& 0.523    &0.622      & 0.752 &0.328\\
TASCNet\cite{li2018learning}  
      &0.559 & -  &-   &0.505       &0.598      & -     & -  \\

UPSNet-M-COCO\cite{xiong2019upsnet}
      &0.618 & 0.813& 0.748& 0.576  &0.648      &0.792  & \textbf{0.390}\\

Panoptic FPN\cite{kirillov2019panoptic}
      &0.612 & 0.809&0.744 & 0.540 &\textbf{0.664}       &0.809  &0.364 \\

DeeperLab\cite{yang2019deeperlab} Xception-71
      &0.565 &-    &-    &-     &-    &-     &-    \\

AdaptIS X101 \cite{sofiiuk2019adaptis}
      &0.620 &-    &-    &0.644  &0.587 &0.792 &0.363 \\
Seamless\cite{porzi2019seamless}
      &0.603 &-    &-    &-     &-    &0.775 &0.336\\
UTIPS\cite{li2020unifying}
      &0.614 &0.811 &0.747 &0.547     &0.663 &0.795 &0.337\\
CASNet(joint)
      &0.590 &0.810 &0.715 &0.478     &0.672 &-     &0.328\\
CASNet(separate)
      &\textbf{0.661} &\textbf{0.833} &\textbf{0.784} &0.536  &\textbf{0.752} &-     &0.358\\

\hline
\end{tabular}
\end{center}

\label{tab:tab_pan}
\end{table*}

\textbf{Experiment analysis.}
We have tried Resnet50, ResNet101, and RetinaNet as the backbone, but under joint training, the semantic head does not output the ideal results as expected. The reason is the vanila backbone configuration is difficult to output competent result on semantic segmentation. Since the common attribute support process heavily relies on the quality of semantic segmentation, this is the reason why we choose HRNet as the backbone, which shows superior accuracy for semantic segmentation.

To verify the effectiveness of the common attribute support strategy, we implement experiment that replaces the predicted semantic segmentation results with the ground truth.
Under this configuration, the panoptic segmentation results are listed in Tab. \ref{tab:ins_gt} and Tab. \ref{tab:pan_gt}. We can see that, if the semantic results are 100\% right, the common attribute head can predict the instances at mAP \(57.4\%\) which is far more than any current algorithm by a large margin. This is also the upper bound of instance segmentation by CASNet. It is also an evidence that the common attribute support method is an effective 
way of identifying instances at pixel level.

\begin{table}
\caption{Instance segmentation results of using semantic ground truth.}
\begin{center}
\begin{tabular}{|l|cc|}
\hline
thing classes      &             AP       &  ${AP_{50}}$\\
\hline\hline
person         &          0.506       &   0.725\\
rider          &          0.663       &   0.778\\
car            &          0.572       &   0.793\\
truck          &          0.560       &   0.603\\
bus            &          0.708       &   0.773\\
train          &          0.635       &   0.645\\
motorcycle     &          0.458       &   0.627\\
bicycle        &          0.488       &   0.705\\
\hline
average        &          0.574       &  0.706\\
\hline
\end{tabular}
\end{center}

\label{tab:ins_gt}
\end{table}

\begin{table}
\caption{The panoptic segmentation results of using semantic ground truth.}
\begin{center}
\begin{tabular}{|c|ccc|}
\hline
Category &All    &Thing    &Stuff\\
\hline\hline
PQ  &0.848    &0.673    &0.974\\
SQ  &0.946    &0.907    &0.974\\
RQ  &0.891    &0.742    &1.000\\
\hline
\end{tabular}
\end{center}

\label{tab:pan_gt}
\end{table}

\section{Conclusion}
An one-stage instance and panoptic segmentation framework Common Attribute Support Network (CASNet) is proposed.
CASNet is inspired from the intuition of each pixel in an instance contains the information of common attribute. 
It works in the fully convolutional way which  can implement training and inference from end to end.
CASNet can get the results without any overlaps and holes between/among instances, which problem exists in most of the current algorithms.
Furthermore, it can be easily extended to panoptic segmentation by adding little computation overhead. 
Through experiments, CASNet achieves comparable results at instance segmentation task and panoptic segmentation on Cityscapes validation split set. 
In the future, more techniques of joint training and light weight backbone need to be explored to get a faster with high prediction accuracy solution.



\bibliographystyle{IEEEtran}
\bibliography{egbib}

\end{document}